%% file: remi.tex
\documentclass[acmtog]{acmart}

\renewcommand\footnotetextcopyrightpermission[1]{} 
\usepackage{url}  
\usepackage{graphicx}  
\usepackage{graphics}
\usepackage{amssymb}
\usepackage{balance}
\usepackage{color}
\usepackage{array}
\usepackage{balance}
\usepackage{booktabs} 
\usepackage[utf8]{inputenc}
\usepackage[ruled,vlined,linesnumbered]{algorithm2e}
\usepackage{amsmath}
\usepackage{algorithmicx}
\usepackage[noend]{algpseudocode}
\usepackage{multirow}
\usepackage{enumitem}

\newcommand{\fact}[3]{$\textit{#2}(\textit{#1}, \textit{#3})$}
\algnewcommand{\IIf}[1]{\algorithmicif\ #1\ \algorithmicthen}
\algnewcommand{\IIIf}[1]{\State\algorithmicif\ #1\ \algorithmicthen}
\algnewcommand{\EndIIf}{\unskip\ \algorithmicend\ \algorithmicif}
\algnewcommand{\WWhile}[1]{\algorithmicwhile\ #1\ \algorithmicdo}

\setcopyright{none}

\begin{document}

\title{REMI: Mining Intuitive Referring Expressions on Knowledge Bases}

\author{Luis Galárraga}
\affiliation{Inria, Univ. Rennes, CNRS, IRISA}
\email{luis.galarraga@inria.fr}

\author{Julien Delaunay}
\affiliation{Univ. Rennes}
\email{juliendelaunay35000@gmail.com}

\author{Jean Louis Dessalles}
\affiliation{Télécom ParisTech}
\email{dessalles@telecom-paristech.fr}


%
%
%
%

\begin{abstract}
A \emph{referring expression} (RE) is a description 
that identifies a set of instances unambiguously. 
Mining REs from data finds applications 
in natural language generation, algorithmic journalism, and data maintenance. Since there may exist
multiple REs for a given set of entities, it is common to focus on the most intuitive ones, i.e., the most concise and informative. 
In this paper we present REMI, a system that can mine intuitive REs on large RDF knowledge bases.
Our experimental evaluation shows that REMI finds REs deemed intuitive 
by users. Moreover we show that REMI is several orders of magnitude faster than
an approach based on inductive logic programming.

\end{abstract}
\maketitle              
\thispagestyle{empty}

\section{Introduction}
\label{sec:introduction}
\input{introduction}
\section{Preliminaries}
\label{sec:preliminaries}
\input{preliminaries}
\section{REMI}
\label{sec:algorithm}
\input{algorithm}

\section{Experimental Evaluation}
\label{sec:experiments}
\input{evaluation}

\section{Related Work}
\label{sec:relatedwork}
\input{relatedwork}

\section{Conclusion and Future Work}
\label{sec:conclusion}
\input{conclusion}

\balance

\bibliographystyle{ACM-Reference-Format}
\bibliography{references2}

\end{document}

%% file: introduction.tex
A \emph{referring expression} (RE) is a description that identifies a set of entities unambiguously. 
For instance, the expression ``x is the capital of France'' 
is an RE for Paris, because no other city holds this title. 
%
The automatic
construction of REs is a central task in natural language generation (NLG).
The goal of NLG is to describe concepts in an accurate and compact manner using structured 
data such as a knowledge base (KB). REs also find applications in 
automatic data summarization, algorithmic journalism, virtual smart assistants, and
KB maintenance, e.g., in query generation.
Quality criteria for REs is context-dependent. For instance, NLG and data summarization aim at 
\emph{intuitive}, i.e., \emph{short} and \emph{informative} descriptions. 
In this vibe, it may be more intuitive to describe Paris as 
``the city of the Eiffel Tower'' than as ``the resting place of Victor Hugo''. Indeed, the world-wide prominence of
the Eiffel Tower makes the first RE more interesting and informative to an average user. 

Some approaches can mine intuitive REs from semantic data~\cite{dale1992generating,reiter92,horacek03,graph-based-res}. 
Conceived at the dawn of the Semantic Web, 
these methods are not suitable for current KBs for three main reasons.
Firstly, they were exclusively designed for NLG, which often focuses on mining REs on scenes\footnote{The exhaustive
description of a place and its objects}. 
Such kind of datasets usually have much fewer predicates and instances than today's KBs. While 
RE mining can be naturally formulated as an inductive logic programming (ILP) task, 
our experimental evaluation shows that RE mining challenges ILP solutions because ILP can lead to 
extremely long rules. Besides, ILP is traditionally not concerned with the intuitiveness of its results.
Secondly, most existing approaches are limited to conjunctive expressions on the attributes of the entities, 
e.g., $\textit{is}(x, \textit{City}) \land \textit{country}(x, \textit{France})$. 
However, our experience with today's KBs suggests that this \emph{state-of-the-art 
language} does not encompass all possible intuitive expressions.
For instance, to describe Johann J. Müller, we could resort to the fact that he was 
the supervisor of the supervisor of Albert Einstein, i.e., 
$\textit{supervisor}(x, y) \land \textit{supervisor}(y, \textit{Einstein})$,
which goes beyond the traditional language bias due to the existentially quantified variable $y$.
Thirdly, state-of-the-art RE miners define intuitiveness for REs in terms of number of atoms. 
In that spirit, the single-atom REs $\textit{capitalOf}(x, \textit{France})$
and $\textit{restingPlaceOf}(x, \textit{V. Hugo})$ are equally concise and desirable as descriptions for Paris, even though the 
latter may not be informative to users outside France. 
We tackle the aforementioned limitations with a solution to mine intuitive REs on
large KBs. How to use such REs is beyond the scope of this work, however
we provide hints about potential use cases. In summary, our contributions are:
\begin{itemize}[leftmargin=*]
 \item We propose a scheme to quantify the intuitiveness 
 of entity descriptions from a KB in number of bits, i.e., we estimate their Kolmogorov complexity~\cite{zellner2001simplicity}.
 \item We study to which extent descriptions of low estimated Kolmogorov complexity are deemed intuitive by users. 
 \item We present REMI, 
 a method to mine intuitive (concise and informative) REs on large KBs. 
 REMI extends the state-of-the-art language bias for REs by allowing additional existentially quantified variables
 as in $\textit{mayor}(x, y)\land\textit{party}(y, \textit{Socialist})$. This design choice increases the chances of finding
 intuitive REs for a set of target entities.
\end{itemize} 

%% file: preliminaries.tex
\subsection{RDF Knowledge Bases}
\label{subsec:kbs}
This work focuses on mining REs on 
RDF\footnote{Resource Description Framework} knowledge bases (KBs). A KB $\mathcal{K}$ is a set of 
assertions in the form of triples $p(s, o)$ (also called facts) with $p\in\mathcal{P}$, $s\in\mathcal{I} \cup \mathcal{B}$, and 
$o \in \mathcal{I} \cup \mathcal{L} \cup \mathcal{B}$. In this formulation, 
$\mathcal{I}$ is a set of entities such as London, 
$\mathcal{P}$ is a set of predicates, e.g., \emph{cityIn}, $\mathcal{L}$ is a set of
literal values such as strings or numbers,  
and $\mathcal{B}$ is a set of blank nodes, i.e., anonymous entities. 
An example of an RDF triple is \fact{London}{cityIn}{UK}. KBs often include assertions such as \fact{UK}{is}{Country} 
that state the class of an entity. Furthermore, for each predicate $p \in \mathcal{P}$ we define its
inverse predicate $p^{-1}$ as the relation consisting of all facts  
$p^{-1}(o, s)$ such that $p(s, o) \in \mathcal{K}$\footnote{To be RDF-compliant,
$p^{-1}$ is defined only for triples with $o \in \mathcal{I} \cup \mathcal{B}$}.  

\subsection{Referring Expressions}
\label{subsec:res}
\subsubsection{Atoms.}
An \emph{atom} $p(X, Y)$ is an expression such that $p$ is a predicate and 
$X$, $Y$ are either variables or constants. 
We say an atom has matches in 
a KB $\mathcal{K}$ 
if there exists a function $\sigma \subset \mathcal{V} \times (\mathcal{I} \cup \mathcal{L} \cup \mathcal{B})$ 
from the variables $\mathcal{V}$ of the atom to constants in the KB 
such that $\mu_{\sigma}(p(X, Y)) \in \mathcal{K}$. The operator $\mu_{\sigma}$ returns a new atom such that the constants
in the input atom are untouched, and 
variables are replaced by their corresponding mappings according to $\sigma$.
We call $\mu_{\sigma}(p(X, Y))$ a bound atom and $\sigma$ a matching assignment.
\subsubsection{Expressions.}
\label{subsubsec:expressions}
We say that two atoms are connected if they share at least one variable argument. 
We define a \emph{subgraph expression} $\rho = \bigwedge_{1\le i \le n}{p_i(X_i, Y_i)}$, rooted at a variable $x$,
as a conjunction of transitively connected atoms such that (1) there is at least an atom that contains $x$ as first
argument\footnote{If $x$ is the second argument as in $p(Y, x)$, we can rewrite the atom as $p^{-1}(x, Y)$} and (2) 
for $n>1$, every pair of atoms is transitively connected via at least another variable besides $x$. Examples of
subgraph expressions rooted at $x$ are:

\begin{enumerate}[leftmargin=*]
 \item $\mathit{cityIn}(x, \mathit{France})$
 \item $\mathit{cityIn}(x, y) \land \mathit{officialLanguage}(y, z) \land \mathit{langFamily}(z, \mathit{Romance})$
 \item $\mathit{cityIn}(x, y) \land \mathit{largestCity}(x, y)$
\end{enumerate}
A subgraph expression $\rho$ has matches in a KB $\mathcal{K}$ 
if there is an assigment $\sigma$ from the variables in the expression to constants in the KB 
such that $\mu_{\sigma}({p_i(X_i, Y_i)}) \in \mathcal{K}$ for $1\leq i \leq n$ in the subgraph expression.
Subgraph expressions are the building blocks of referring expressions, thus 
we define an \emph{expression} $e = \bigwedge_{1 \leq j \leq m} \rho_j$ as a conjunction of 
subgraph expressions rooted at the same variable $x$
such that the expressions have only $x$---called the root variable---as common variable. 
An expression $e$ has matches in a KB if there is an assigment $\sigma$ that yields matches for every subgraph expression 
of $e$.
Finally, we say an expression $e$ with root variable $x$ is a \emph{referring expression} (RE) 
for a set of target entities $T \subseteq \mathcal{I}$ in a KB 
$\mathcal{K}$ iff the following two conditions are met:
\begin{enumerate}[leftmargin=*]
 \item $\forall t \in T : \exists \sigma : (x \mapsto t) \in \sigma$, i.e., 
 for every target entity $t$, there exists a matching assignment $\sigma$ in $\mathcal{K}$ 
 that binds the root variable $x$ to $t$.
 \item $\nexists\sigma', t' : (x \mapsto t') \in \sigma' \land \;t' \not\in T$, 
 in other words, no matching assignment binds the root variable to entities outside the set $T$ of target entities.
\end{enumerate}
\noindent For example, consider a KB $\mathcal{K}$ with accurate and complete information about countries and languages, as well as 
the following expression $e$ rooted at $x$ and consisting of two subgraph expressions:
\begin{small}
\[e = \textit{in}(x, \textit{South America}) \land \textit{officialLanguage}(x, y) \land \textit{langFamily}(y, \textit{Germanic}) \]
\end{small}
\noindent We say that $e$ is an RE for entities $T = \{\mathit{Guyana},\mathit{Suriname}\}$ in $\mathcal{K}$ 
because matching assignments can only bind $x$ to these two countries.

%% file: algorithm.tex
Given an RDF KB $\mathcal{K}$ and a set of target entities
$T$, REMI returns the most intuitive RE---a conjunction of subgraph expressions---that describes
unambiguously the input entities in $\mathcal{K}$. 
We define intuitiveness for REs as a trade-off between compactness and informativeness, thus intuitive REs should be 
\emph{simple}, i.e., they should be concise and resort to concepts that users are likely to know and understand. We first quantify intuitiveness for REs
in number of bits in Section~\ref{subsec:complexityfunction}.
We then elaborate on REMI's language bias and algorithm 
in Sections~\ref{subsec:languagebias} and~\ref{subsec:algorithm} respectively.

\subsection{Quantifying intuitiveness}
\label{subsec:complexityfunction}
There may be multiple ways to describe a set of entities unambiguously. For instance, 
$\textit{capitalOf}(x, \textit{France})$ and $\textit{birthPlaceOf}(x, \textit{Voltaire})$ are both REs for Paris.
Our goal is therefore to quantify the \emph{intuitiveness} of such expressions without human intervention.
We say that an RE $e$ is more intuitive than an RE $e'$, if $C(e) < C(e')$, where $C$ 
denotes the Kolmogorov complexity; so we define intuitiveness as the inverse of complexity. 
The Kolmogorov complexity $C(e)$ of a string $e$ (e.g., an expression) is a measure of the absolute 
amount of information conveyed by $e$ and is defined
as the length in bits of $e$'s shortest effective binary description~\cite{zellner2001simplicity}.
If $e_b$ denotes such binary description and $M$ is the program that can \emph{decode} $e_b$ into $e$,  
$C(e) = l(e_b) + l(M)$ where $l(\cdot)$ denotes length in bits.
The exact calculation of $C(e)$
requires us to know the shortest way to compress $e$ as well as the most compact program $M$ for decompression.
Due to the intractability of $C$'s calculation, applications can only \emph{approximate} it by means of potentially 
suboptimal encodings and programs, that is: 
$$C(e) \le \hat{C}(e) = l(\hat{e_b}) + l(\hat{M})$$ 
Once we fix a compression/decompression scheme $\hat{M}$, applications only need to worry 
about the term $l(\hat{e_b})$. 
To account for the dimension of informativeness, our proposed encoding builds upon the observation that intuitive expressions resort to prominent predicates and entities. 
For example, it is very natural and informative 
to describe Paris as the capital of France, because the notion of capital is well understood and France is a very prominent entity. 
In contrast, it would be more complex to describe Paris in terms of its twin cities, because this concept is less 
prominent than the concept of capital city. 
In other words, prominent predicates and entities are informative because they 
relate to concepts that users often recall. We can thus devise a code for predicates and entities by constructing
a \emph{ranking by prominence}. The code for a predicate $p$ (entity $I$) is the binary representation of its position $k$
in the ranking and the length of its code is $\mathit{log}_2(k)$. In this way, prominent concepts
are encoded with fewer bits.
We now define the estimated Kolmogorov complexity $\hat{C}$ of a single-atom subgraph expression $p(x, I)$ as:
\begin{small}
\[\hat{C}(p(x, I)) = l(p_b) + l(I_b\;|\;p) \]
\end{small}
In the formula $p_b = k(p)$, where $k(p)$ is the position of predicate $p$ in the ranking of predicates.
It follows that $l(p_b) = \mathit{log}_2(k(p))$.
The term $l(I_b\;|\;p) = \mathit{log}_2(k(I\;|\;p))$ accounts
for the chain rule of the Kolmogorov complexity. It measures the conditional complexity of $I$ given predicate $p$.
It is calculated as the logarithm of $I$'s rank in the ranking of 
objects of predicate $p$. For instance, let us assume that $p$ is the predicate \emph{city mayor}. The chain rule models 
the fact that once the concept of mayor has been conveyed, the context becomes narrower and the user needs
to discriminate among fewer concepts, in this example, only city mayors. The chain rule also applies to 
subgraph expressions with multiple atoms. For instance, the complexity of  
$\rho = \textit{mayor}(x, y) \land \textit{party(y, \textit{Socialist})}$ corresponds to the following formula:
\[
\begin{aligned}
\hat{C}(\rho) ={} & l(\textit{mayor}_b) + l(\textit{party}(y, z)_b \: | \: \textit{mayor(x, y)}) \: + \\ 
& l(\textit{Socialist}_b \: |\: \textit{mayor}(x, y) \land \textit{party}(y, z)) 
\end{aligned}
\]
We highlight that the code for \emph{party} must account for the fact that this predicate appears in 
a first-to-second-argument join with the predicate \emph{mayor}. Hence, the second term in the sum amounts to 
$\mathit{log}_2(k(\textit{party}(y, z)\: | \: \textit{mayor(x, y)}))$, the log$_2$ of the rank of \emph{party} among those predicates that allow for 
first-to-second-argument joins with \emph{mayor} in the KB. Likewise, the complexity of the Socialist party in the third term is derived
from a ranking of all the parties that have mayors among their members, i.e., all the bindings for the variable $z$ 
in $\textit{mayor}(x, y) \land \textit{party}(y, z)$.
If a city can be unambiguously described by its non-prominent mayor $I$, 
it may be simpler to instead omit the person and describe her by paying the price of an additional 
predicate and a well-known party. 

We can now estimate the Kolgomorov complexity of a referring expression $e = \bigwedge_{1\le i \le m}{\rho_i}$ as the 
sum of the complexities of the individual subgraph expressions, namely as $\hat{C}(e) = \sum_{1\le i \le m}{\hat{C}(\rho_i)}$. 
Consider as an example the following RE:
$$e = \mathit{in}(x, \mathit{S. America})\land \mathit{officialLang}(x, y) \land \mathit{langFamily}(y, \mathit{Germanic})$$
It follows that we can calculate $\hat{C}(e)$ as: 
\[
\begin{aligned}
\hat{C}(e) ={} & \hat{C}(\mathit{in}(x, \mathit{S. America})) \;+ \\ 
& \hat{C}(\mathit{officialLang}(x, y) \land \mathit{langFamily}(y, \mathit{Germanic})) 
\end{aligned}
\]
It is vital to remark, however, that this formula makes a simplification. Consider the RE 
$\textit{officialLang}(x, IT) \land \textit{officialLang}(x, DE)$ for Switzeland. 
While $\hat{C}$ adds the complexity of the predicate \emph{officialLang} twice, an optimal code would count the predicate
once and encode its multiplicity. In fact, this optimal code 
would be applied to every common sub-path with multiplicity in the list of subgraph expressions. 
This fact worsens the quality of $\hat{C}$ as an approximation of $C$ for such kind of expressions, however 
it is not a problem in our setting as long as we use $\hat{C}$ for comparison purposes.

Lastly, we discuss how to rank concepts by prominence. Wikipedia-based KBs
provide information about the hyperlink structure of the entity pages, thus
one alternative is to use the Wikipedia page rank (PR). The downside of this metric is that it is undefined for predicates.
A second alternative is frequency, i.e., number of mentions of a concept.
Frequency could be measured in the KB or extracted from exogenous sources
such as a crawl of the Web or a search engine. Even though search engines may quantify prominence 
more accurately (by providing real-time feedback and 
leveraging circumstantiality), 
we show that endogenous sources are good enough for this goal.
In line with other works that quantify popularity for concepts in KBs~\cite{linksum}, we 
use (i) the number of facts where a concept occurs in the KB (\emph{fr}), and (ii) the concept's page rank in Wikipedia (\emph{pr}).
We denote the resulting complexity measures using these prominence metrics 
by $\hat{C}_{\mathit{fr}}$ and $\hat{C}_{\mathit{pr}}$ respectively. We use \emph{fr} whenever \emph{pr} is undefined.
\subsection{Language Bias}
\label{subsec:languagebias}
Most approaches for RE mining define REs as conjunctions of atoms with bound objects,
thus we call this language bias, \emph{the state-of-the-art language bias}.
REMI extends this language by allowing atoms with
additional existentially quantified variables. 
This design decision allows us to replace tail entities with high Kolmogorov complexity 
with entities that are more prominent and hence more intuitive. 
For instance, consider the RE $\textit{supervisorOf}(x, \textit{Alfred Kleiner})$ for
Johann J. Müller. Saying that he was the supervisor of A. Kleiner may not
say much to an arbitrary user. By allowing an additional variable, we can consider the expression
``he was the supervisor of the supervisor of Albert Einstein'', namely
$\textit{supervisorOf}(x, y) \land \textit{supervisorOf}(y, \textit{A. Einstein})$. 
We highlight that Einstein is \emph{simpler} to describe than Kleiner, 
which makes the second expression, albeit longer, more informative and overall more intuitive than the first one. 
This shows that in the presence of irrelevant object entities, further atoms may help increase intuitiveness. 
Nevertheless, in the general case longer expressions tend to be more complex. 
This phenomenon becomes more palpable when the additional atoms 
do not describe the root variable as in 
$\textit{speaks}(x, y) \land \textit{family}(y, z) \land \textit{superfamily}(z, \textit{Italic})$ (``she 
speaks a language in a subfamily of the Italic languages''). 
This expression introduces two additional variables that 
turn comprehension and translation to natural language more effortful. 
Besides, further atoms and specially additional variables can dramatically increase the size of the search space of REs, 
which is exponential in the number of possible subgraph expressions.
Our observations reveal, e.g., that a second additional variable increases by more than 270\% the number
of subgraph expressions that REMI must handle in DBpedia. 
Conversely, increasing the number 
of atoms from 2 to 3 while keeping only one additional variable, leads to an increase of 40\%. 
Based on all these observations, we restrict REMI's language
bias to subgraph expressions with at most one additional variable and 3 atoms. 
The 3-atom constraint
goes in line with rule mining approaches on large KBs~\cite{amieplus}. 
This decision disqualifies our
last example, but still allows expressions such as $\textit{bornIn}(x, y) \land \textit{livedIn}(x, y) \land 
\textit{diedIn}(x, y)$ (she was born, lived and died in the same place).
Table~\ref{tab:language} summarizes REMI's language
of subgraph expressions.

\begin{table}
  \centering
  \begin{tabular}{ l  l  }
    \toprule
    1 atom & $p_0(x, I_0)$ \\
    Path & $p_0(x, y) \land p_1(y, I_1)$ \\
    Path + star & $p_0(x, y) \land p_1(y, I_1) \land p_2(y, I_2) $ \\
    2 closed atoms\; & $p_0(x, y) \land p_1(x, y)$ \\ 
    3 closed atoms\;  & $p_0(x, y) \land p_1(x, y) \land p_2(x, y)$ \\ \bottomrule
  \end{tabular}
  \caption{REMI's subgraph expressions.}
  \label{tab:language}
\end{table}
\begin{figure}
\includegraphics[width=0.45\textwidth]{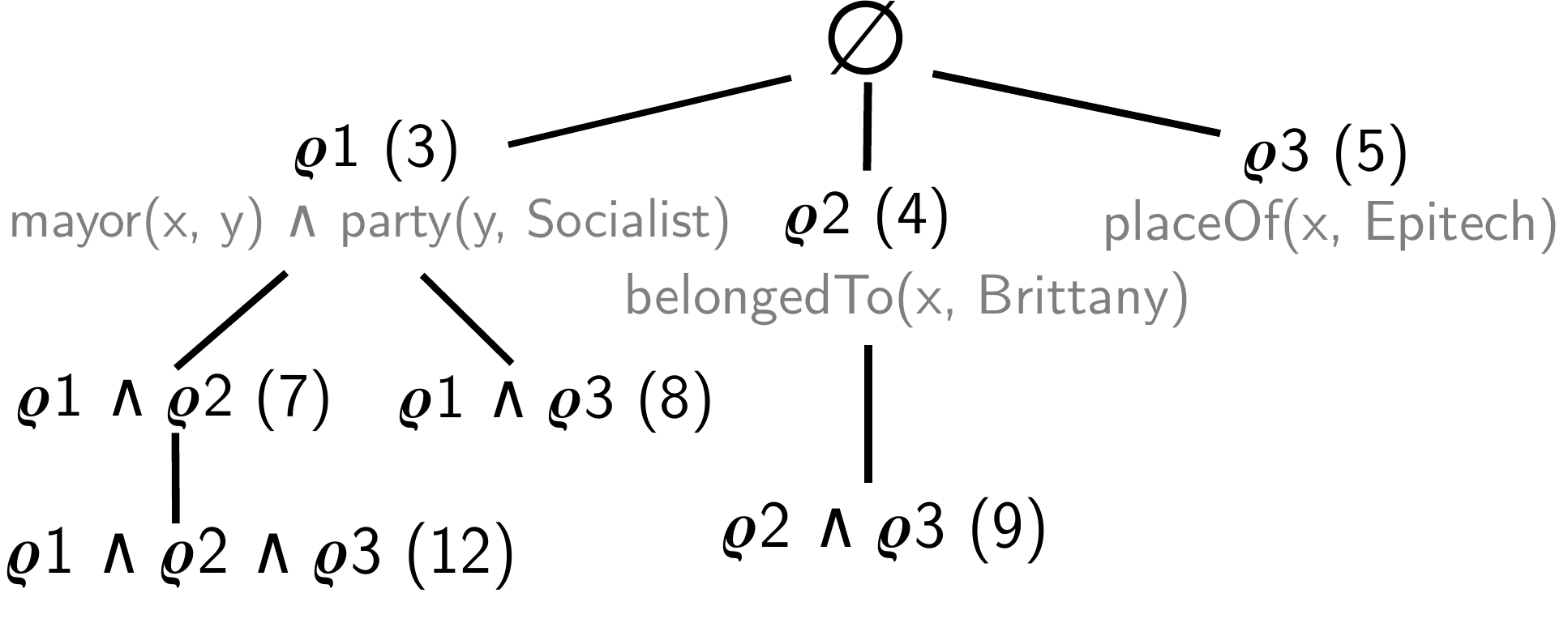}
\caption{Search space example.}
\label{fig:tree}
\end{figure}

\subsection{Algorithm}
\label{subsec:algorithm}
REMI implements a depth-first search (DFS\footnote{DFS approaches are preferred over BFS (breadth-first search) due to their smaller
memory footprint})
on conjunctions of the subgraph expressions common
to \textbf{all} the target entities.
Assume that the KB knows only three common subgraph expressions $\rho_1$, $\rho_2$, and $\rho_3$ for the entities 
\emph{Rennes} and \emph{Nantes}, such that $\hat{C}(\rho_1) \leq \hat{C}(\rho_2) \leq \hat{C}(\rho_3)$.
The tree in Figure~\ref{fig:tree} illustrates the search space for our example. 
Each node in the tree is an expression,
i.e., a conjunction of subgraph expressions and its complexity $\hat{C}$ is in parentheses.
When visiting a node, DFS must test whether the corresponding expression is an RE, i.e., 
whether the expression identifies exclusively the target entities. If the test fails, the strategy should move to the node's first child.
If the test succeeds, DFS must verify whether the expression is less complex
than the least complex RE seen so far. If it is the case, this RE should be remembered, and DFS can prune
the search space by backtracking. 
To see why, imagine that $\rho_1 \land \rho_2$ in Figure~\ref{fig:tree} is an RE.
In this case, all REs prefixed with this expression (the node's descendants) 
will also be REs. However, all these REs have higher values for $\hat{C}$, i.e., they are more complex.
This means that we can stop descending in the tree pruning the node 
$\rho_1 \land \rho_2 \land \rho_3$ in Figure~\ref{fig:tree}. We call this step a \emph{pruning by depth}.
We can do further pruning if we leverage the order of the entities. In our example, if 
$\rho_1 \land \rho_2$ is an RE, any expression prefixed with $\rho_1 \land \rho_{i}$ 
for $i>2$ must be more complex and can be therefore skipped. 
We call this a \emph{side pruning}. All these ideas are formalized by Algorithm~\ref{alg:remi} that takes as input a
KB $\mathcal{K}$ as well as the entities to describe, and returns an RE of minimal complexity according to
$\hat{C}$.
Line 1 calculates all subgraph expressions in REMI's language bias (Figure~\ref{tab:language})
that are common to the target entities. 
They are then sorted by increasing complexity and put in a priority queue (line 2). The depth-first exploration is 
performed in lines 4-8. At each iteration, the least complex subgraph expression $\rho$ is dequeued (line 5) 
and sent to the subroutine \emph{DFS-REMI} (line 6) with the rest of the queue.
This subroutine returns the most intuitive RE $e'$ prefixed with $\rho$. 
If $\hat{C}(e')$ is smaller than the complexity of the best
solution found so far (line 7), we remember it\footnote{We define $\hat{C}(\top) = \infty$}.
If \emph{DFS-REMI} returns an empty expression, we can conclude that there is no RE for the target entities $T$ (line 8).
To see why, recall that DFS will, in the worst case, combine $\rho$ 
with all remaining expressions $\rho'$ that are more complex.
If none of such combinations is an RE, there is no solution for $T$ in $\mathcal{K}$.
\begin{algorithm}
\begin{small}
\caption{REMI}
\label{alg:remi}
\KwIn{a KB: $\mathcal{K} \subseteq (\mathcal{I} \cup \mathcal{B}) \times \mathcal{P} \times (\mathcal{I} \cup \mathcal{L} \cup \mathcal{B})$, 
the target entities: $T \subset \mathcal{I}$}
\KwOut{an RE of minimal complexity: $e$}
    $G := \bigcap_{t \in T}{\textit{subgraphs-expressions}(t)}$ \\
    create priority queue from $G$ in ascending order by $\hat{C}$ \\    
    $\textit{e} := \top$ \\
    
    \While{$|G| > 0$}{
	$\rho := G.\textit{dequeue}()$ \\
	$e' := $\emph{DFS-REMI}($\rho$, $G$, $T$, $\mathcal{K}$) \\
	\IIf{$\hat{C}(e') < \hat{C}(e)$}{ $e := e'$ \\} 
	\IIf{$e = \top$}{ \Return $\top$ \\} 

    }
    \Return $e$
\end{small}
\end{algorithm}

\noindent We now sketch how to calculate the subgraph expressions of an entity $t$ (line 1). 
Contrary to \emph{DFS-REMI}, the routine $\textit{subgraphs-expressions}$ carries out a 
breadth-first search. Starting with 
atomic expressions of the form $p_0(x, I_0)$ (where $x$ binds to $t$), the routine derives all two-atom expressions, 
namely paths of the form $p_0(x, y) \land p_1(y, I_1)$ and closed conjunctions 
$p_0(x, y) \land p_1(x, y)$. 
The two-atom paths are extended with atoms of the form $p_2(y, I_2)$ to produce the path+star combinations,
whereas the closed conjunctions are used to derive the closed expressions of three atoms (see Table~\ref{tab:language}).

The subroutine \emph{DFS-REMI} is detailed in Algorithm~\ref{alg:dfs-remi}, which 
takes as input a subgraph expression $\rho$, the priority queue of subgraph expressions 
$G$ (without $\rho$), the target entities $T$ and a KB $\mathcal{K}$.
We use a stack initialized
with the empty subgraph expression $\top$ in order to traverse the search space in a depth-first manner (line 1). 
Each iteration pushes a subgraph expression to the stack, starting with $\rho$ (line 3).
The conjunction of the elements of the stack defines an expression $e'$ that 
is evaluated on the KB to test if it is an RE for the target entities (lines 4-5). If $e'$  
is the least complex RE seen so far, the algorithm remembers it (line 6).
Adding more expressions to $e'$ can only increase
its complexity, hence line 7 performs a pruning by depth, so that all descendants of $e'$ are abandoned. 
Line 8 backtracks anew to achieve a side pruning. 
If backtracking leads
to an empty stack, \emph{DFS-REMI} cannot do better, and can thus return $e'$ (line 9). 
\begin{algorithm}
\caption{DFS-REMI}
\label{alg:dfs-remi}
\KwIn{a subgraph expression: $\rho$, priority queue: $G$, target entities: $T$, a KB: $\mathcal{K}$}
\KwOut{an RE of minimal complexity prefixed in: $e$}
     $\textit{S} := \{\top \}; \; e := \top; \; G' := \{\rho \} \cup G$ \\
     \ForEach{$\rho' \in G'$}{
	$\textit{S} := S \cup \{ \rho' \} $ \\
	$e' := \bigwedge_{\hat{\rho} \in S}{\hat{\rho}}$ \\
	\If{ $e'(\mathcal{K}) = T$}{
	  \IIf{$\hat{C}(e') < \hat{C}(e)$}{ $e := e'$ \\ }
	  $S.\mathit{pop}()$ \\
	  $S.\mathit{pop}()$ \\
	  \IIf {$S = \emptyset$}{ \Return $e$ \\ }
	}
    }
    \Return $e$
\end{algorithm}
\subsection{Parallel REMI} 
We can parallelize Algorithm~\ref{alg:remi} if we allow multiple threads to concurrently dequeue elements from
the priority queue of subgraph expressions and explore the subtrees rooted at those elements independently. 
This implies to execute the loop in lines 4-8 in parallel.
This new strategy, called P-REMI, preserves the logic of REMI with three differences.
First, the least complex solution $e$ can be read and written by all threads.
Second, if a thread found no solution from its exploration rooted at subgraph expression $\rho_i$, 
it must signal all the other threads rooted at subgraph expressions
$\rho_j$ ($j > i$) to stop. For instance, if a thread finished its exploration rooted at $\rho_1$ in Fig.~\ref{fig:tree},
any exploration rooted at $\rho_2$ or $\rho_3$ is superfluous as it covers expressions
that are less specific than those rooted at $\rho_1$.
Third, before
testing if an expression is an RE, each thread should verify whether there is already a solution $e$
of lower complexity. If so, the thread can backtrack until reaching
a node of even lower complexity than $e$. 
Since these differences mostly affect
the logic of \emph{DFS-REMI}, we detail a new procedure called \emph{P-DFS-REMI} in Algorithm~\ref{alg:p-dfs-remi}.
The new routine has the same signature as \emph{DFS-REMI} plus a reference to the best solution $e$. 
Lines 1 and 2 initialize the stack and create a new priority queue $G'$ from 
the original one. The DFS exploration starts in line 3. The first task of \emph{P-DFS-REMI} is to
backtrack iteratively while the expression
represented by the stack is less complex than the best solution $e$ (line 6). 
If \emph{P-DFS-REMI} backtracked to the root node, it means the algorithm cannot find a better solution from now
on, and can return $e$ (line 7). If backtracking did not remove any expression from the stack (check in line 8), 
\emph{P-DFS-REMI} proceeds exactly as its sequential counterpart \emph{DFS-REMI} (lines 9-14). 
Conversely, if the stack was pruned by the loop in line 6, P-DFS-REMI starts a new iteration since the
corresponding expression in the resulting stack must have been tested in a previous iteration. 
For proper implementation the access to $e$ must be synchronized among the different threads. 
\begin{algorithm}
\caption{P-DFS-REMI}
\label{alg:p-dfs-remi}
\KwIn{a subgraph expression: $\rho$, priority queue of subgraph expressions: $G$,
the target entities: $T$, $e$: best solution found so far, 
a KB: $\mathcal{K}$}
\KwOut{an RE of minimal complexity: $e$}
     $S := \{\top\} $ \\
     $G' := \{ \rho \} \cup  G$ \\
     \While{$|G'| > 0$}{
	$\rho' := G'.\textit{dequeue}()$ \\
	$\textit{S} := S \cup \{ \rho' \} $ \\
	\WWhile{$|S| > 1 \land \hat{C}(\bigwedge_{\hat{\rho} \in S}{\hat{\rho}}) \ge \hat{C}(e)$}{
	  $S.\textit{pop}()$
	}
	
	\IIf{$S = \{ \top \}$}{ \Return $e$ }
	
	\If{$S.\mathit{peek()} = \rho' $}{
	  $e' := \bigwedge_{\hat{\rho} \in S}{\hat{\rho}}$ \\	
	  \If{ $e'(\mathcal{K}) = T$}{
	    \IIf{$\hat{C}(e') < \hat{C}(e)$}{ $e := e'$ \\ }
	    
	    $S.\mathit{pop}()$ \\
	    $S.\mathit{pop}()$ \\
	    \IIf {$S = \emptyset$}{ \Return $e$ \\ }
	  }
	}
    }
    \Return $e$
\label{subsec:premi}
\end{algorithm}
\subsection{Implementation}\label{subsec:implementation}
\subsubsection{Data storage.} We store the KB in a single file using the HDT~\cite{hdt} format. 
HDT is a binary compressed format, conceived for fast data transfer, that offers 
reasonable performance for search and browse operations without prior decompression.
HDT libraries support only the retrieval of bindings for atoms $p(X, Y)$, leaving 
the execution of additional query operators to upper layers. We used the Apache Jena framework\footnote{\url{https://jena.apache.org/}} (version 3.7)
as access layer. 
\subsubsection{Algorithms.}
\label{subsubsec:algorithms}
REMI and P-REMI are implemented in Java 8. Their runtime is dominated
by two phases: (1) the construction of the priority queue of subgraph expressions (line 2 in Alg.~\ref{alg:remi}), and (2)
the DFS exploration (lines 4-8). The first phase is computationally expensive
because it requires the calculation of $\hat{C}$ on large sets 
of subgraph expressions, leading to the execution of expensive queries on the KB.
To alleviate this fact, we parallelized the construction and sorting of the queue
and applied a series of pruning heuristics.
First, the routine $\textit{subgraphs-expressions}$ ignores expressions
of the form $p(x, B)$ with $B \in \mathcal{B}$, since blank nodes are by conception irrelevant entities.
However, the routine always derives paths that ``hide'' blank nodes, that is, 
$p(x, y) \land p'(y, I)$ (such that $y$ binds to $B$ and $I \in \mathcal{I}$) is always considered.
Conversely, we do not derive multi-atom subgraph expressions
from atoms with object entities among the 5\% most prominent entities. 
For example, we do not explore extensions of $\textit{capitalOf}(x, \mathit{Germany})$
such as $\textit{capitalOf}(x, y) \land \textit{locatedIn}(y, \emph{Europe})$, because 
the complexity of the additional atom will likely be higher than the complexity of a simple entity such as \emph{Germany}. 
Finally, REMI requires the execution of the same queries multiple times, thus query results are cached
in a least-recently-used fashion. 
\subsubsection{Complexity function.}
The calculation of $\hat{C}$ requires the construction of multiple rankings on prominence
for concepts in the KB. For example,
to calculate   
$\hat{C}(\textit{capital}(x, \textit{Paris}))$, we need the rank $k(\textit{capital})$
among all predicates, as well as the rank $k(\textit{Paris} \;|\; \textit{capital})$ 
among all capital cities. 
Even though predicates are always evaluated against the same ranking,
an entity may rank differently depending on the context.  
We could precompute $k(I \;|\; p)$ for every $I$, $p$ in the KB, however we can leverage the
correlation between prominence and rank to reduce the amount of stored information. 
It has been empirically shown that the frequency of 
terms in textual corpora follows a power-law distribution~\cite{IR}. 
If $\mathit{fr}(k)$ is the frequency of the k$^\text{th}$
most frequent term in a corpus, $\mathit{fr}(k) \approx \hat{\beta} k^{-\hat{\alpha}}$ for some constants $\hat{\alpha}, \hat{\beta} > 0$. 
If we treat all the facts $p(s, o) \in \mathcal{K}$ with the same predicate $p$ as a corpus, we can 
estimate the number of bits of an entity given $p$ from its conditional frequency $\mathit{fr}(I \;|\;p) = |I: \exists s : p(s, I) \in \mathcal{K}| $ as follows:
\begin{small}
\begin{equation}\label{eq:powerlaw}
\mathit{fr}(I \;|\; p) \approx \hat{\beta} \times k(I \;|\; p)^{-\hat{\alpha}} \therefore \; \mathit{log}_2(k(I \;|\; p)) \approx -\alpha \mathit{log}_2(\mathit{fr}(I \;|\; p)) + \beta
\end{equation}
\end{small}
\noindent We can thus learn the 
coefficients $\alpha$ and $\beta$ that map frequency in the KB to complexity in bits.  
While this still requires us to precompute the conditional rankings, Equation~\ref{eq:powerlaw} allows us to ``compress'' them
as a collection of pairs of coefficients (one per predicate).
Our results on two KBs confirm the linear correlation between the logarithms of rank and frequency,
since the fitted functions exhibit an average $R^2$ measure of 0.85 in DBpedia
and 0.88 in Wikidata\footnote{Values closer to 1 denote a good fit.}.
Likewise, this power-law correlation extrapolates to the Wikipedia page rank, which reveals an 
average $R^2$ of 0.91 in DBpedia. 



%% file: evaluation.tex
We evaluated REMI along two dimensions: output quality  
and runtime performance.
The evaluation was conducted on  
DBpedia~\cite{dbpedia} and Wikidata~\cite{wikidata}. 
For DBpedia (v. 2016-10) we used the files instance types, mapping-based objects,
and literals. The resulting dataset amounts to 42.07M facts and 1951 predicates. 
For Wikidata we used the dump provided in~\cite{completeness-paper} that contains 15.9M facts and 752 predicates.
For both KBs we materialized the inverse facts $p^{-1}(o,s)$ ---such that $p(s,o)\in \mathcal{K}$--- 
for all objects $o$ among the top 1\% most frequent entities. 


\subsection{Qualitative Evaluation}
\label{subsec:qualityevaluation}
We carried out four rounds of experiments in order to evaluate the intuitiveness of REMI's descriptions.
These experiments comprise three user studies, and an evaluation with a benchmark for entity summarization.
The cohort for the user studies consisted mainly of computer science students, researchers, and university staff.
It also included some of their friends and family members. 
\subsubsection{Evaluation of $\hat{C}$.} Recall that subgraph expressions are the building blocks of REs, 
thus intuitive REs should make use of simple (concise and informative) components.
In order to measure whether function $\hat{C}$ captures users' notion of intuitiveness,
our first study asked participants
to rank by simplicity five subgraph expressions from 24 sets of 
DBpedia entities. 
The expressions were obtained from the common subgraph expressions ranked by Alg.~\ref{alg:remi} in
line 2 using $\hat{C}$, and include (i) the top 3 as well as a baseline defined by (ii) the worst ranked, and (iii) a random
subgraph expression.  
We manually translated the subgraph expressions to natural language statements in the shortest 
possible way by using the textual descriptions (predicate \emph{rdfs:label}) of the concepts when available.
The entity sets (of sizes 1 to 3) were randomly sampled from the 5\% most frequent entities in four classes: 
Person, Settlement, Album $\cup$ Film, and Organization. Most of these entities are tail entities, however, 
looking at the top of the frequency ranking ensures that the entities have enough
subgraph expressions to rank. We show the results in Table~\ref{tab:qualitative} for our two variants of $\hat{C}$ using 
\emph{fr} and \emph{pr} as prominence measure. 
We observe that precision@1 is low for both versions of $\hat{C}$. 
The reason behind this result is that people usually 
deem the predicate type the simplest whereas REMI often ranks it second or third (16 times for $\hat{C}_{\mathit{fr}}$). This shows
the need of special treatment for the \emph{type} predicate as suggested by~\cite{reiter92}. 
Nevertheless, the high values for the other metrics 
show a positive correlation between the preferences of the users and the function $\hat{C}$. 
In 88\% of the cases, the three simplest
subgraph expressions according to $\hat{C}$ are among the three simplest ones according to users. 
\begin{table*}
\centering
\begin{tabular}{>{\centering\arraybackslash}m{2.2cm}>{\centering\arraybackslash}m{2.2cm}>{\centering\arraybackslash}m{1.5cm}>{\centering\arraybackslash}m{1.8cm}>{\centering\arraybackslash}m{1.8cm}>{\centering\arraybackslash}m{1.8cm}}
  \toprule
   \textbf{metric} 		& \textbf{\# responses} & \textbf{p@1} 	 		& \textbf{p@2}      & \textbf{p@3}  \\
      $\hat{C}_{\mathit{fr}}$   &	44     		   &   0.38$\pm$0.42	 	&  0.66$\pm$0.18    &  0.88$\pm$0.09  \\ 
      $\hat{C}_{\mathit{pr}}$   &	48      	   &   0.43$\pm$0.42 	 	&  0.53$\pm$0.25    &  0.72$\pm$0.16  \\ \bottomrule
\end{tabular}
\caption{Average precision@k ($k\in \{1, 2, 3\}$) and standard deviation for the ranking of subgraph expressions computed by two variants of $\hat{C}$ in DBpedia}
\label{tab:qualitative}
\end{table*}

\subsubsection{Evaluation of REMI's output.} The second study requested users to rank by simplicity
the answer of REMI and a baseline consisting of other 
REs encountered during search space traversal. 
We hand-picked 20 sets of prominent entities of the same DBpedia classes used to evaluate $\hat{C}$, and provided 
3 to 5 solutions for each set including the most intuitive RE according to REMI. 
The entities were hand-picked to guarantee the existence of at least two solutions that are 
not too similar to each other, i.e., they are not proper subsets of other solutions. 
Based on our findings with the first study and the higher complexity of this task, 
we provided the type of the entities in the question statement and used \emph{fr} as notion of prominence. 
We report an average MAP (mean average precision) 
of 0.64$\pm$0.17 for this task on 51 answers when we assume
REMI's solution as the only relevant answer. To give the reader an idea of this result, we recall that
a MAP of 1 denotes full agreement between REMI and the users, while a MAP of 0.5 means that REMI's solution is always
among the user's top 2 answers. Furthermore, we remark that in 6 out of the 20 
studied cases, REMI reported the same solution with $\hat{C}_{\mathit{fr}}$ and $\hat{C}_{\mathit{pr}}$ as intuitiveness metric. 
When we asked users to choose the simplest RE between the answers of the two variants of REMI, 
59\% of the users on average voted for the solution provided by $\hat{C}_{\mathit{fr}}$.
\subsubsection{User's perceived quality and lessons learned.}
In order to measure the perceived quality of the reported REs, we requested users to grade the 
\emph{interestingness} of 35 Wikidata REs in a scale from 1 to 5, where 5 means the user deems 
the description interesting based on her personal judgment. 
The entities were taken from the top 7 of the frequency ranking for the 
classes Company, City, Film, Human, and Movie. Our results on 86 answers exhibit an average score of 2.65$\pm$0.71, with 11
descriptions scoring at least 3. 
During the exchanges with the participants, some of them made explicit their preference for short 
but at the same time informative REs. The latter dimension is related 
to the notions of pertinence of concepts and narrative interest. For instance, users deemed uniformative and 
scored badly (1.45/5)
REMI's portrayal of Neil Amstrong as someone ``whose place of burial is a part of the earth'' (in allusion to the fact 
that he was buried in the Atlantic Ocean). When asked to choose between
the REs $\mathit{country}(x, \mathit{N.\text{ }Zealand}) \land \mathit{actor}(x, \mathit{C.\text{}Lee})$ and
$\mathit{country}(x, \mathit{N.\text{ }Zealand}) \land \mathit{actor}(x, y) \land \mathit{religion}(x, \mathit{Buddhism})$ 
for two movies, 95\% of the users preferred the first one\footnote{They correspond to the answers of REMI with the two variants
of $\hat{C}$}. 
Both REs had more or less the same length when translated to natural language, 
but the second one conveys less information and resorts to a domain-unrelated entity (i.e., Buddhism). 
These observations suggest that prominence captures the notion of simplicity, but it does not always
accurately model the dimension of informativeness. While 
these examples might discourage the use of existential variables in descriptions, 
we remark that users also liked REs such as 
$\mathit{in}(x, \mathit{Brittany}) \land \mathit{mayor}(x, y) \land \mathit{party}(y, \mathit{Socialist})$
for Rennes and Nantes, or $\mathit{actor}(x, y) \land \mathit{leader}(y, \mathit{Pisa})$ for the Italian movie ``Altri templi'', 
as they deemed the first one quite pertinent, and the second one narratively interesting. 
Other well-ranked descriptions include ``the CEOs is Andrej Babiš, the Prime Minister of the Czech Republic''
(scored 3.97/5) for Agrofert in Wikidata, ``she died of aplastic anemia'' for Marie Curie, 
and ``they were both places of the Inca Civil War'' for Ecuador and Peru (the two latter in DBpedia).
Finally, we highlight the impact of noise and incompleteness in the accuracy and informativeness of the solutions. For instance,
REMI cannot describe France as the country with capital Paris, because Paris is also the capital of the former 
Kingdom of France in DBpedia.


\subsubsection{Evaluation on benchmark for entity summarization.} Despite being different tasks, RE mining and entity 
summarization in KBs are related problems (see Section~\ref{sec:relatedwork} for a discussion). For this reason, we evaluated REMI on the 
gold standard
used for the evaluation of FACES~\cite{faces} and LinkSUM~\cite{linksum}, two state-of-the-art approaches 
for entity summarization on RDF KBs. The gold standard consists of sets of reference summaries of 5 and 10 attributes
for 80 prominent hand-picked  entities from DBpedia. 
The entity summaries were manually constructed by 7 experts in semantic Web, and consist of pairs 
predicate-object chosen from DBpedia with
diversity, prominence, and uniqueness as selection criteria.
We ran REMI with the state-of-the-art language bias, and excluded 
the subgraph expressions with the predicate \emph{rdf:type} and the inverse predicates $p^{-1}$
to make our results compliant with the language of the summaries.
We compare the reference summaries with the top 5 and top 10 most intuitive 
subgraph expressions (single atoms in this case) according to
$\hat{C}_{\mathit{fr}}$ and $\hat{C}_{\mathit{pr}}$ in Table~\ref{tab:faces}.

\begin{table*}
\centering
\begin{tabular}{p{2.2cm}>{\centering\arraybackslash}m{2.4cm}>{\centering\arraybackslash}m{2.4cm}>{\centering\arraybackslash}m{2.4cm}>{\centering\arraybackslash}m{2.4cm}}
  \toprule
    \multirow{2}{*}{\textbf{Method}}	& \multicolumn{2}{c}{\textbf{top 5}} 					& \multicolumn{2}{c}{\textbf{top 10}}   		\\ 
					&    \textbf{quality PO}		& \textbf{quality O}  		&    	\textbf{quality PO}		& \textbf{quality O}			\\	
    FACES 				&	0.93$\pm$0.54			&	1.66$\pm$0.57		& 	2.92$\pm$0.94  			& 4.33$\pm$1.01	\\
    LinkSUM     			&	1.20$\pm$0.60			&	1.89$\pm$0.55		& 	3.20$\pm$0.87  			& 4.82$\pm$1.06	 \\ 
    REMI $\hat{C}_{\mathit{fr}}$ 	&	0.68$\pm$0.18			&	1.31$\pm$0.27		&	2.26$\pm$0.34			& 3.70$\pm$0.46	\\ 
    REMI $\hat{C}_{\mathit{pr}}$ 	&	0.73$\pm$0.13			&	1.21$\pm$0.29		&	2.24$\pm$0.46			& 3.75$\pm$0.23	\\ \bottomrule    
\end{tabular} 
\caption{REMI's top 5 and top 10 subgraph expressions compared with solutions for entity summarization in RDF KBs.}
\label{tab:faces} 
\end{table*} 
\noindent Quality is defined in~\cite{faces} as the average overlap between the reported and the gold standard summaries. 
This overlap can be calculated at the level of the object entities (O) or the pairs predicate-object (PO).
Even though the quality of REMI's summaries exhibits a lower variability than other approaches, its average quality 
is generally lower. This happens because entity summarization approaches 
optimize for a different objective. Besides (non-strict) unambiguity, and the use of popular concepts, these approaches 
optimize for diversity too. This implies that among multiple semantically close subgraph expressions, summaries must choose only 
one. We remark that such a constraint makes sense in a setting when a description may consist of 10 atoms, 
however it may be too restrictive for settings such as NLG or query generation where compacteness matters. 
Finally, we highlight that if we merge the top-10 gold-standard summaries, the REs reported by REMI yield averages of 
0.53, 0.62, and 0.31 for the P, O, and PO precisions when using $\hat{C}_{\mathit{fr}}$ as prominence metric, i.e., 
62\% of the RE's used object entities
that appear in the summaries. The values for $\hat{C}_{\mathit{pr}}$ are slightly worse, except for the PO precision (0.38).
Most of the REs reported by REMI consisted of a single atom.

\subsection{Runtime Evaluation}
\label{subsec:runtimeevaluation}
\subsubsection{Opponent.}
RE mining can be conceptually formulated as an ILP task. Hence,
we compare the runtime of REMI and a state-of-the-art parallel ILP system designed for large KBs, namely AMIE+~\cite{amieplus}.
We chose AMIE+ over other solutions, because it allows us to mine rules of arbitrary
length out of the box.
AMIE+ mines Horn rules of the form $p(X, Y) \Leftarrow \bigwedge_{1 \leq i \leq n}{p_i(X_i, Y_i)}$,
such as $\mathit{speaks}(x, \mathit{English}) \Leftarrow \mathit{livesIn}(x, \mathit{UK})$,
on RDF KBs. 
We call the left-hand side the \emph{head} and the right-hand side the \emph{body} of the rule. 
AMIE+ focuses on closed rules, i.e., rules where all variables appear in at least two atoms. 
The system explores the space of closed rules in a breadth-first search manner and reports those 
above given thresholds on support and confidence. 
The \emph{support} of a rule is the number of cases where the rule predicts a fact $p(s, o) \in \mathcal{K}$.
If we normalize this measure by the total number of predictions made by the rule, we obtain its \emph{confidence}.
RE mining for a target entity set $T$ is equivalent to rule mining with AMIE+, if we instruct the system to find rules of the form 
$\psi(x, \mathit{True}) \Leftarrow \bigwedge_{1 \leq i \leq n}{p_i(X_i, Y_i)}$, where $\psi$ is a surrogate predicate
with facts $\psi(t, \mathit{True})$ for all $t \in T$. 
In this case, the body of the rule becomes our RE.
We observe, however, that AMIE+ cannot capture REMI's language bias exactly.
This happens because AMIE's language is defined in terms of a maximum number of
atoms $l$, whereas REMI allows an arbitrary number of multi-atom subgraph expressions. 
Since most of REMI's descriptions are not longer than 3 atoms,
we set $l=4$ for AMIE+.
We also set thresholds of $|T|$ and 1.0 for support and confidence respectively.
This is because an RE should predict the exact set of target entities, neither subsets nor supersets. 
AMIE+ does not define a complexity score for rules and outputs all REs for the target entities, thus
we use $\hat{C}_{\textit{fr}}$ to rank AMIE's output and return
the least complex RE.
\begin{table*}
\centering
\begin{tabular}{c|ccccc|ccccc}
  \toprule
    \multirow{2}{*}{Language}	& \multicolumn{5}{c|}{DBpedia} 															& \multicolumn{5}{c}{Wikidata}   \\ \cline{2-11}
				&    \#solutions	&amie+ 					&  remi				& p-remi   			& speed-up 		& \#sol.	& amie+					&  remi						& p-remi  				& speed-up 	\\	\hline	
    Standard 			& 	63		&97.4k{\color{red}$^{8}$}    		&  10.3k{\color{red}$^1$}	& 576  				& 13.5kx, 2.44x		& 44		& 115.5k{\color{red}$^{15}$}		&  1.06k					& 76.2					& 142kx, 4.7x	\\
    REMI's     			& 	65		&508.2k{\color{red}$^{68}$}   		&  66.5k{\color{red}$^{8}$}	& 28.9k	  			& 5218x, 21.4x		& 44		& 608.3k{\color{red}$^{60}$}		&  21.7k					& 33.8k					& 6476x, 7.1x	\\ \bottomrule
\end{tabular} \vspace{0.2cm}
\caption{REMI's runtime performance (in seconds) on DBpedia and Wikidata. The column speed-up denotes the average speed-up 
of P-REMI w.r.t. AMIE+ and REMI.}
\label{tab:runtime}
\end{table*} 
\subsubsection{Results.}
We compared the runtimes of REMI and AMIE+ 
on a server with 48 cores (Intel Xeon E2650 v4), 
192GB of RAM\footnote{AMIE assumes the entire KB fits to main memory}, and 1.2T of disk space (10K SAS).
We tested the systems on 100 sets of DBpedia and Wikidata entities  
taken from the same classes used in the qualitative evaluation. The sets were randomly chosen so that  
they consist of 1, 2, and 3 entities of the same class 
in proportions of 50\%, 30\%, and 20\%. We chose sets of at most 3 entities
because small sets translate into more subgraph expressions leading to 
more challenging settings.  
%
We mined REs for those sets of entities according to (i) the standard 
language of conjunctive bound atoms, and (ii) REMI's language of conjunctions of 
subgraph expressions. 
We show the total runtime among all sets for AMIE+ and REMI in Table~\ref{tab:runtime}. For each group of entities, we set a
timeout of 2 hours.
The values in red account for the number of timeouts, thus cells with red superscripts 
define runtime lower bounds. We observe that AMIE+ already timed out 23 times with the state-of-the-art language. 
In particular, AMIE+ is optimized for rules without constant arguments in atoms, such as $\mathit{livesIn}(x, y) \Leftarrow \mathit{citizenOf}(x, y)$, thus its
performance is heavily affected when bound variables are allowed in atoms~\cite{amieplus}.
In contrast REMI and P-REMI are on average 3 and 4 orders of magnitude (up to 142k times) faster than AMIE+ in this language 
thanks to the tailored pruning heuristics detailed in Section~\ref{subsec:implementation}. 
In the worst case REMI was confronted with a space of 62 subgraph expressions for the state-of-the-art language bias. 
For REMI's 
language bias, however, this number increased to 25.2k, which is challenging even for REMI (8 timeouts in total). 
Despite this boost in complexity, multithreading makes it manageable: 
P-REMI can be \emph{at least} 4.7x on average faster than REMI for the extended language bias 
and \emph{at least} 21x faster for the state-of-the-art language, even though PREMI's total time on Wikidata
is higher than REMI's for the extended language bias. 
The latter phenomenon is caused by the high variance of the speed-up, which ranges 
from 0.003x---for small search spaces where the overhead of multithreading is overwhelming---to 197x in Wikidata. Extending the language bias also 
increases the time to sort the subgraph expressions (line 2 in Alg.~\ref{alg:remi}), which
jumps from 0.39\% to 9.1\% for P-REMI in DBpedia.
Finally, we observe that the extended language bias slightly increases the chances of finding a solution (column \#solutions in Table~\ref{tab:runtime}) in DBpedia.
We observe this phenomenon among the sets with more than one entity.

%% file: relatedwork.tex
Mining REs on structured data is a central task in natural language generation (NLG). 
Systems such as Epicure~\cite{epicure} and IDAS~\cite{idas} provided descriptions in natural language
based on REs mined on domain-specific KBs ---equipment parts and recipes respectively.
NLG methods consider criteria such as brevity, context, user's prior knowledge, lexical preference, 
and psychological factors when producing simple and informative REs~\cite{pechmann1989incremental}.
The \emph{full brevity algorithm}~\cite{dale1992generating}, based on breadth-first search,
is among the first approaches to mine REs on semantic data.
This method mines short REs consisting of conjunctions of bound atoms, which we call the standard language bias. 
Nonetheless, the results of~\cite{dale1992generating} are not always intuitive, because 
it disregards factors such as user's prior knowledge and lexical preference.
The incremental approach proposed in~\cite{reiter92} took these criteria into account by 
modeling user's knowledge as Boolean metatags on facts, 
and lexical preference as a manually-constructed ranking of predicates.
While this solution produces more intuitive REs, providing these metadata can be tedious for large 
KBs with thousands of classes and predicates. In REMI, 
the complexity ranking for subgraph expressions
captures both lexical preference and user's knowledge automatically to a certain degree, as subgraph expressions 
with familiar concepts are ranked higher.  
Other approaches~\cite{horacek03} allow for REs with disjunctions, e.g., 
$\mathit{officialLang}(x, \mathit{ES}) \lor \mathit{officialLang}(x, \mathit{FR})$. 
Albeit more expressive, the language of REs with disjunctions is in general 
more difficult to interpret. REMI proposes conjunctive expressions with existentially
quantified variables instead. These are more expressive than standard REs, and
can be intuitive as shown in Section~\ref{subsec:qualityevaluation}.
The work in~\cite{graph-based-res} finds standard conjunctive REs
in knowledge graphs by means of a branch and bound algorithm in combination with
several cost functions. Such functions can optimize for different aspects such as compactness. 
Unlike REMI, this approach was conceived to mine REs on scenes, hence it does not scale to large KBs.
The largest graph tested had 256 vertices and 1.7K edges. 

Other works focus on expressions that are similar to REs. Maverick~\cite{maverick}, e.g., mines 
exceptional facts on KBs. Given an entity such as Hillary Clinton and a context, e.g., 
``candidates to the US presidential election'', Maverick will report the fact that she is a female, 
as that makes her exceptional in the context. Unlike REMI, Maverick does not find REs in a strict sense; instead
it reports descriptions that are \emph{rare} among the entities in the context. Approaches for entity 
summarization (ES), such as~\cite{linksum,faces}, construct informative summaries of entities from a KB by considering 
groups of attributes
that optimize for uniqueness, diversity, and prominence. Albeit related to RE mining, ES 
is concerned neither with compacteness nor with strict unambiguity: ES approaches usually take the size 
of the summary as input, and do not guarantee that the resulting summary is a strict RE. 
Besides, ES approaches has not been defined for sets of entities to the best of our knowledge.
All these approaches mine expressions in the standard language bias.

%% file: conclusion.tex
In this work we have presented REMI, a method to mine intuitive referring expressions on large RDF KBs. 
REMI relies on the observation that frequent concepts 
are more intuitive to users, and leverages this fact to quantify the simplicity (intuitiveness) of descriptions in bits.
We have not targeted a particular 
application in this work, instead we aim at paving the way towards the automatic generation of descriptions in
large KBs. 
Our results show that (1) real-time RE generation is often possible in large KBs and (2) a KB-based frequency ranking can 
provide intuitive descriptions despite the noise in KBs. While this latter
factor impedes the fully automatic generation of intuitive REs for NLG purposes, 
our descriptions are applicable to scenarios such as
computer-aided journalism and query generation in KBs.
As future work we aim to investigate if external sources---such as the ranking provided by 
a search engine or external localized corpora---can yield even more intuitive REs that model users' background more accurately. 
We also envision to relax the unambiguity
constraint to mine REs with exceptions, and study more accurate models  
for the dimensions of informativeness and semantic relatedness for the concepts used in REs. 
We provide the source code of REMI as well as the experimental data at \url{https://gitlab.inria.fr/lgalarra/remi}.